\documentclass[10pt,twocolumn,letterpaper]{article}

\usepackage[pagenumbers]{cvpr} 

\usepackage{times}
\usepackage{epsfig}
\usepackage{graphicx}
\usepackage{amsmath}
\usepackage{amssymb}
\usepackage{paralist}
\usepackage{float}
\usepackage{stfloats}
\usepackage{booktabs}
\usepackage{multicol}
\usepackage{multirow}
\usepackage{bm}
\usepackage{bbm}
\usepackage{diagbox}
\usepackage{makecell}
\usepackage{algorithm}  
\usepackage{algorithmic} 
\usepackage{enumitem} 
\usepackage{tabularx} 
\usepackage{threeparttable} 
\usepackage{listings} 
\usepackage{color, colortbl}
\usepackage[accsupp]{axessibility}
\newcommand{\gr}{\rowcolor[gray]{.95}} 
\newcommand{\good}{\rowcolor[RGB]{234,243,233}}
\definecolor{cvprblue}{rgb}{0.21,0.49,0.74}
\usepackage[pagebackref,breaklinks,colorlinks,allcolors=cvprblue]{hyperref}

\def\ours{KnowVal}
\newcommand{\figref}[1]{Figure~\ref{#1}}%
\newcommand{\tabref}[1]{Table~\ref{#1}}%

\renewcommand{\eqref}[1]{Eq.~(\ref{#1})}

\newcommand{\add}[1]{{\scriptsize\color{magenta}{#1}}}

\def\paperID{12999} 
\def\confName{CVPR}
\def\confYear{2026}

\title{\ours: A Knowledge-Augmented and Value-Guided \\Autonomous Driving System}

\author{Zhongyu Xia$^1$\quad Wenhao Chen$^1$\quad Yongtao Wang$^1$\thanks{Corresponding author.}\quad Ming-Hsuan Yang$^2$
\\
$^1$Wangxuan Institute of Computer Technology, Peking University
\\
$^2$University of California, Merced
\\
{\tt\small \{xiazhongyu,wyt\}@pku.edu.cn
\quad \tt\small wenhaochen@stu.pku.edu.cn}
\quad \tt\small mhyang@ucmerced.edu
}

\begin{document}
\maketitle
\begin{abstract}

Visual–language reasoning, driving knowledge, and value alignment are essential for advanced autonomous driving systems. However, existing approaches largely rely on data-driven learning, making it difficult to capture the complex logic underlying decision-making through imitation or limited reinforcement rewards.
To address this, we propose \ours, a new autonomous driving system that enables visual–language reasoning through the synergistic integration of open-world perception and knowledge retrieval. Specifically, we construct a comprehensive driving knowledge graph that encodes traffic laws, defensive driving principles, and ethical norms, complemented by an efficient LLM-based retrieval mechanism tailored for driving scenarios. Furthermore, we develop a human-preference dataset and train a Value Model to guide interpretable, value-aligned trajectory assessment.
Experimental results show that our method substantially improves planning performance while remaining compatible with existing architectures. 
Notably, \ours~achieves the lowest collision rate on nuScenes and state-of-the-art results on Bench2Drive and NVISIM.

\end{abstract}    
\section{Introduction}
\label{sec:intro}

Autonomous driving, a key application of AI, serves as a comprehensive testbed for 3D perception, prediction, reasoning, and planning capabilities.
Current state-of-the-art autonomous driving systems mainly follow two paradigms:
(a) End-to-end autonomous driving, which leverages 3D feature extraction to directly decode planning trajectories; and
(b) Vision–Language–Action (VLA) models, which integrate perception and control through multi-modal large language models.
However, an ideal autonomous driving system must operate in open, dynamic, and highly uncertain environments, requiring decisions that are safe, reliable, and explainable while adhering to human social values and norms, which poses significant challenges to existing paradigms.

\begin{figure}[t]
    \centering
    \includegraphics[width=\columnwidth]{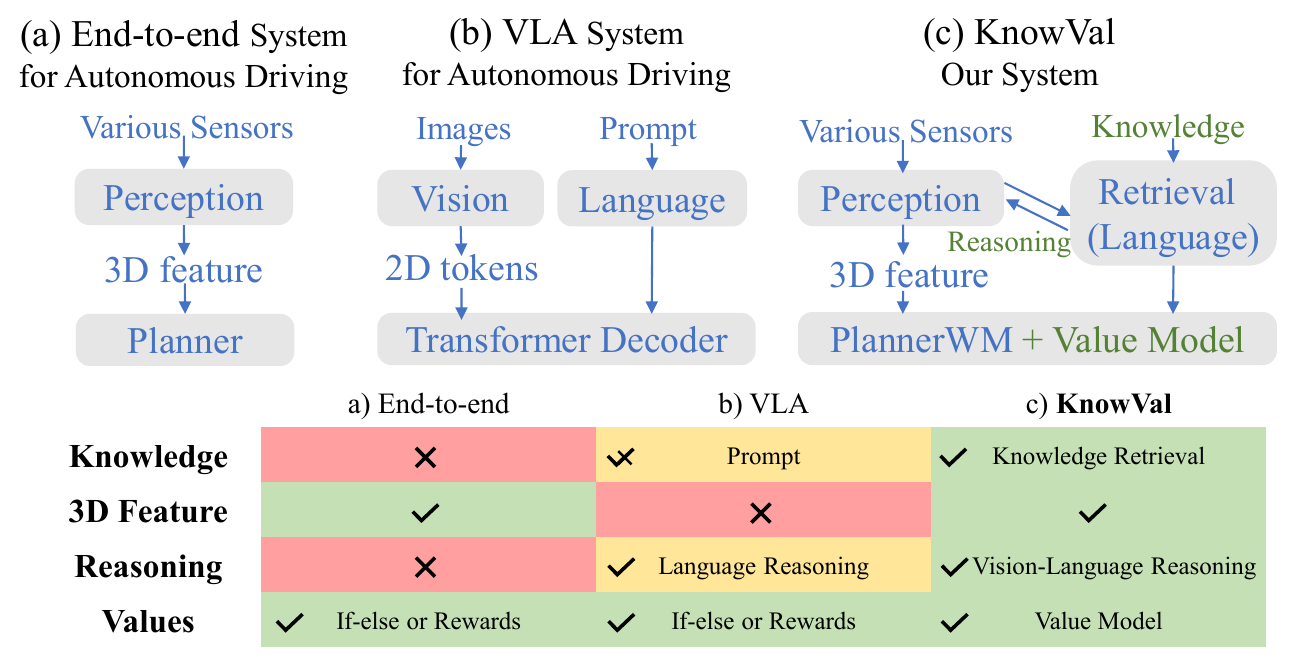}
    \vspace{-20pt}
  \caption{\textbf{Comparison of End-to-End, VLA, and our \textsc{\ours}.} 
    Our 3D vision system enables visual–language reasoning through the mutual guidance of perception and retrieval, understands \textit{law and morality} via knowledge retrieval, and integrates \textit{values} through a dedicated value model.}
    \label{fig:intro}
    \vspace{-20pt}
\end{figure}

The first challenge lies in enabling effective interaction between vision and language to achieve genuine visual–language reasoning. Existing end-to-end autonomous driving models lack language-grounded reasoning capabilities, while VLA models typically confine reasoning to chain-of-thought or purely linguistic processes, without allowing the reasoning outcomes to influence perception.
As illustrated in Figure~\ref{fig:intro}, we propose a visual–language reasoning system in which perception and retrieval mutually guide each other: the model retrieves relevant knowledge from open-world perception, and when retrieval identifies missing information, it prompts the perception module to refine its observations accordingly.

The second challenge lies in the absence of an interpretable and structured knowledge base. Existing end-to-end and VLA models primarily rely on data-driven learning, attempting to infer complex decision logic from the limited variability of human behavior—an inherently difficult task. Some approaches introduce handcrafted conditional rules or reinforcement learning rewards, but these manually designed components cover only narrow scenarios and fail to generalize to complex, dynamic environments.
In contrast, autonomous driving already has a rich and authoritative source of knowledge: traffic laws and regulations, which provide a comprehensive foundation for safe and lawful driving. Furthermore, abundant online resources, such as instructional driving videos, offer valuable insights into driving techniques and principles. Hence, we fully exploit these resources to construct a knowledge graph of traffic laws, moral principles, and defensive driving behaviors, enabling dynamic knowledge retrieval via perception queries. This allows the model to interpretably reason about traffic laws, moral principles, and defensive driving behaviors.

The third challenge concerns the autonomous driving system's worldview and values. Recent advances highlight the importance of a world model—a module that predicts future world states from current observations and actions—reflecting the predictive and reflective abilities of the autonomous driving systems. However, possessing such a worldview (\textit{i.e.}, world model) alone is insufficient for effective decision-making: after forecasting possible futures, the system must also judge whether those outcomes are desirable, requiring a dedicated value assessment mechanism.
Existing paradigms often approximate this ability through data-driven learning or handcrafted rules, both of which limit generalization and interpretability. To address this, we introduce a Value Model grounded in retrieved knowledge and trained on a human-preference dataset, enabling value-aligned and interpretable trajectory evaluation.

To sum up, we propose a novel autonomous driving system in this work, in which perception and knowledge retrieval mutually guide each other, supplying richer contextual information to the decision-making module that integrates both a world model and a value model.
The key contributions of this work are:

\begin{compactitem}
    \item 
    We propose a new autonomous driving system, \ours, that enables visual–language reasoning through the synergistic interaction between perception and knowledge retrieval.
    \item 
    We construct a comprehensive driving knowledge graph encompassing traffic laws, defensive driving principles, and moral considerations, and develop an efficient LLM-based retrieval mechanism for it.
    \item 
    We design a planner that integrates a world model for future-state prediction and a Value Model for outcome evaluation, enabling value-aligned decision-making. 
    A human-preference dataset is also curated to train the Value Model.
    \item 
    Extensive experiments show that \ours~maintains strong compatibility with existing methods, and achieves the lowest collision rate on nuScenes and state-of-the-art performance on Bench2Drive and NVISIM.
\end{compactitem}

\section{Related Work}
\label{sec:related}

\noindent \textbf{Atonomous Driving Systems.}
The prevailing paradigms for autonomous driving systems can be broadly categorized into three classes: the traditional modular pipeline, the end-to-end approach, and the Vision–Language–Action (VLA) model.
Several models have been benchmarked on real-world datasets such as nuScenes~\cite{nuScenes} and NAVSIM~\cite{navsim}. ST-P3~\cite{hu2022st} first integrates perception, prediction, and planning into an end-to-end framework. UniAD~\cite{UniAD} follows a BEV-based feature extraction strategy similar to BEVFormer~\cite{BEVFormer}, with intricate feature propagation across task-specific decoders. VAD~\cite{jiang2023vad} adopts a fully vectorized scene representation for trajectory planning, while PARA-Drive~\cite{weng2024drive} further investigates multi-task supervision and module interdependencies within end-to-end architectures.
GenAD~\cite{zheng2024genad} utilizes BEV representations to obtain sparse detection and mapping features for planning, whereas SparseDrive~\cite{sun2025sparsedrive} extracts such features directly from image inputs to mitigate cumulative errors and improve prediction accuracy. Recently, MomAD~\cite{song2025don} introduces momentum-aware planning for enhanced temporal consistency, and BridgeAD~\cite{zhang2025bridging} refines planning results by leveraging historical prediction and trajectory information. Finally, DiffusionDrive~\cite{liao2025diffusiondrive} employs a truncated diffusion model to generate diverse trajectory candidates, while HENet++~\cite{henetpp} jointly extracts foreground instance and dense panoramic features using attention mechanisms to improve prediction and planning.

With the advancement of 3D physical simulators, many recent works~\cite{UniAD, jiang2023vad, liao2025diffusiondrive} leverage synthetic environments such as CARLA~\cite{dosovitskiy2017carla} and Bench2Drive~\cite{jia2024bench2drive} for training and evaluation. Despite this progress, these methods still largely rely on imitation learning paradigms.
TCP~\cite{wu2022trajectory} introduces a dual-branch architecture that integrates trajectory and control signals to enhance planning accuracy. ThinkTwice~\cite{jia2023think} proposes a two-stage planner that first generates candidate trajectories and then refines them based on scene context, partially reflecting the concept of a world model. DriveAdapter~\cite{jia2023driveadapter} utilizes an expert planner for knowledge distillation, improving planning robustness. AD-MLP~\cite{zhai2023rethinking} extracts priors from historical trajectories using multilayer perceptrons, while DriveTransformer~\cite{jia2025drivetransformer} adopts parallel queries for sparse interactions with sensor data in streaming multi-task learning.
With the emergence of Vision–Language Models (VLMs), several methods have adapted them for driving command generation, among which ORION~\cite{fu2025orion} is representative. SimLingo~\cite{renz2025simlingo} further aligns natural-language prompts (e.g., “adjust speed,” “change lane”) with corresponding action sequences, advancing interpretability and language-conditioned control.

However, these approaches lack vision–language reasoning, explicit knowledge utilization, and value-based judgment.
Our proposed system overcomes these limitations through perception-guided knowledge retrieval and knowledge-guided value assessment, enabling more interpretable and value-aligned decision-making.


\begin{figure*}[htbp]
    \centering
    \includegraphics[width=.95\textwidth]{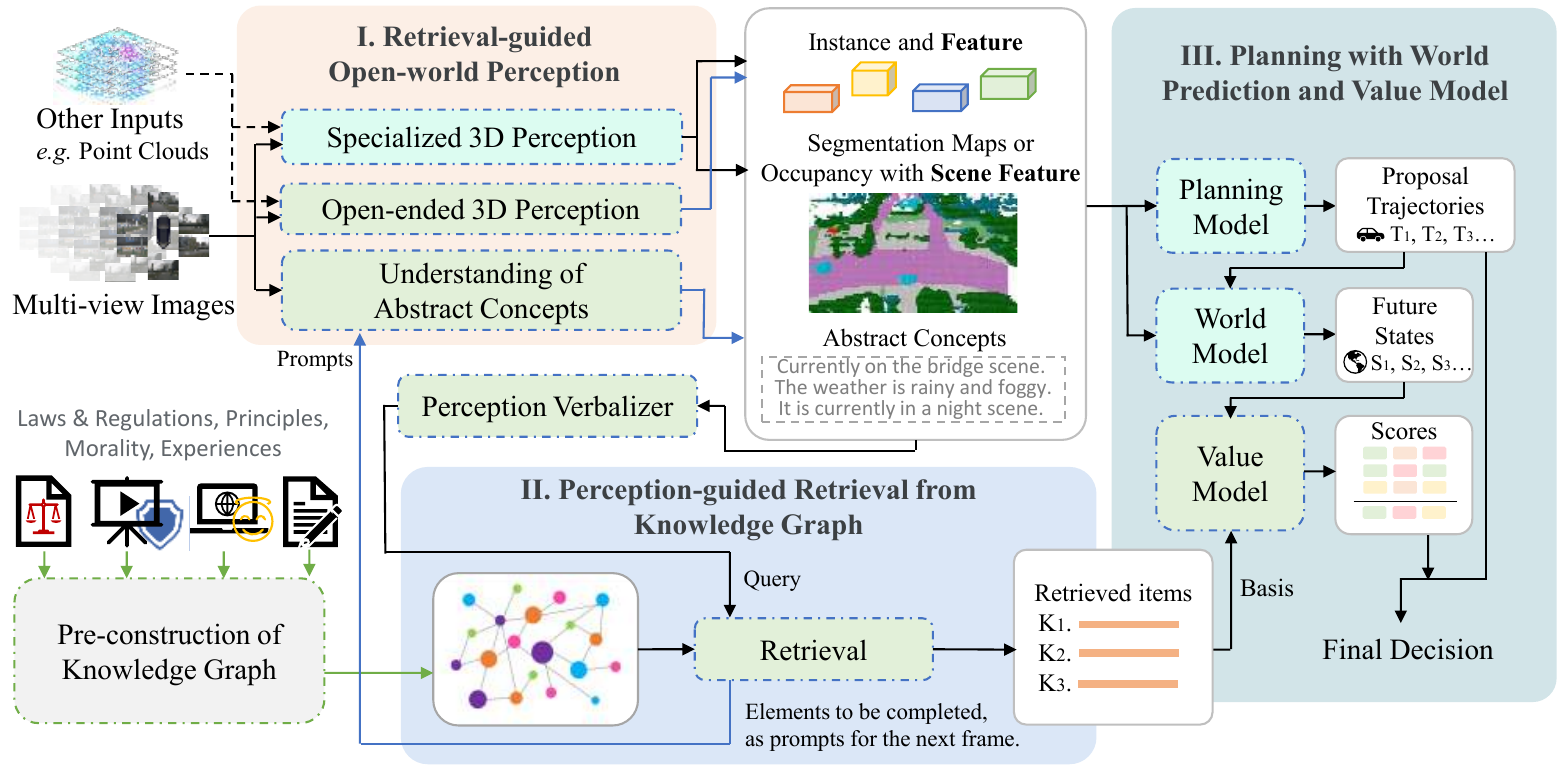}
    \vspace{-8pt}
\caption{\textbf{Overall architecture of \textsc{\ours}.} 
\textsc{KnowVal} first performs \textit{retrieval-guided open-world perception} to extract instance-, panoramic-, and concept-level information. 
It then conducts \textit{perception-guided retrieval} from the knowledge graph to obtain relevant knowledge. 
Finally, decision-making is achieved through integrated planning, world prediction, and value assessment. 
\textsc{\ours} is compatible with existing methods. The Specialized Perception, Planning Model, and World Model are obtained by modifying and fine-tuning existing methods.
Feature propagation across modules ensures comprehensive information flow throughout the system.}
    \label{fig:method}
    \vspace{-15pt}
\end{figure*}

\noindent \textbf{Retrieval Method.}
%
%
To align systems with complex knowledge principles, integrating external knowledge is essential.
Early retrieval-based methods, such as~\cite{ma2023query}, adopt a simple linear pipeline—indexing documents, retrieving the most similar chunks to a query, and generating responses based on them.
The introduction of Retrieval-Augmented Generation (RAG)~\cite{lewis2020retrieval} improves this process by incorporating pre-retrieval and post-retrieval stages to enhance contextual relevance and generation quality.
Modular RAG~\cite{yu2022generate, shao2023enhancing} extends this idea by decomposing the linear RAG pipeline into a flexible architecture with specialized modules arranged in dynamic configurations.
GraphRAG~\cite{edge2024local} constructs a summarized knowledge graph over text corpora, providing global semantic structure and enabling reasoning beyond vector search.
More recently, LightRAG~\cite{guo2024lightrag} introduces a dual-level retrieval mechanism over a graph-enhanced index that supports incremental updates, allowing efficient and adaptive RAG without expensive re-indexing.

However, the aforementioned retrieval-augmented methods, though effective for language models, cannot be directly applied to autonomous driving systems. They inevitably paraphrase and compress retrieved knowledge, which can introduce hallucinations and distort factual content. Moreover, these approaches are primarily designed for chat-based AI, where retrieval is driven by user-issued textual prompts.
To meet the unique demands of autonomous driving, we propose a perception-guided retrieval mechanism that grounds knowledge access in visual perception rather than user input. In addition, we refine the retrieval process to ensure that retrieved entries remain faithful to the original texts, thereby preserving factual accuracy and reliability in downstream reasoning and decision-making.

 

\section{Method}
\label{sec:method}

\subsection{Reasoning between Perception and Retrieval}

As illustrated in~\figref{sec:method}, \ours~comprises three main components:
(i) retrieval-guided open-world perception,
(ii) perception-guided retrieval from the knowledge graph, and
(iii) planning with world prediction and value assessment.
\ours~achieves visual–language reasoning through the mutual guidance between perception and retrieval, thereby enriching both perceptual understanding and knowledge utilization. Information from these two modules is subsequently propagated to the planning component as structured representations and feature embeddings, ensuring comprehensive decision-making while maintaining an end-to-end architecture.

\noindent \textbf{Retrieval-guided Open-world Perception.}
Open-world perception in \ours~encompasses three key capabilities.
The first is Specialized Perception, a standard component in most end-to-end autonomous driving models, which enables the recognition and localization of common semantic categories such as vehicles, pedestrians, and drivable areas.
The second is Open-ended 3D Perception, which identifies and localizes long-tail or uncommon objects (e.g., fire trucks, standing water) without requiring explicit user prompts. This capability is implemented using VL-SAMv2~\cite{lin2025vl} and OpenAD~\cite{xia2024openad}.
The third is Abstract Concept Understanding, which captures scene-level or contextual attributes beyond direct object perception—such as determining whether the environment is a bridge or tunnel, or whether it is daytime or nighttime—achieved via the VLM within VL-SAMv2~\cite{lin2025vl}.
Additionally, this module supports retrieval guidance: the retrieval component from the previous timestep identifies elements requiring further perception, which are then passed to the VLM for targeted processing.

\noindent \textbf{Perception-guided retrieval from knowledge graph.}
We collect traffic laws, regulations, and driver interview transcripts, and employ multimodal large language models (MLLMs) to extract experiential knowledge, ethical considerations, and defensive driving principles from online video sources.
Subsequently, we utilize LLMs to construct a knowledge graph, as described in Section~\ref{sss:kg}.
During inference, a Perception Verbalizer transforms structured perception outputs into query texts, which are then used by an LLM-based retrieval module to obtain knowledge entries ranked in descending order of relevance.
Simultaneously, the retrieval module identifies elements requiring additional perception, which are forwarded to the perception component in the next timestep for refinement.
The retrieval process is further detailed in Section~\ref{sss:retrieval}.

\begin{figure*}[ht]
    \centering
    \includegraphics[width=.9\textwidth]{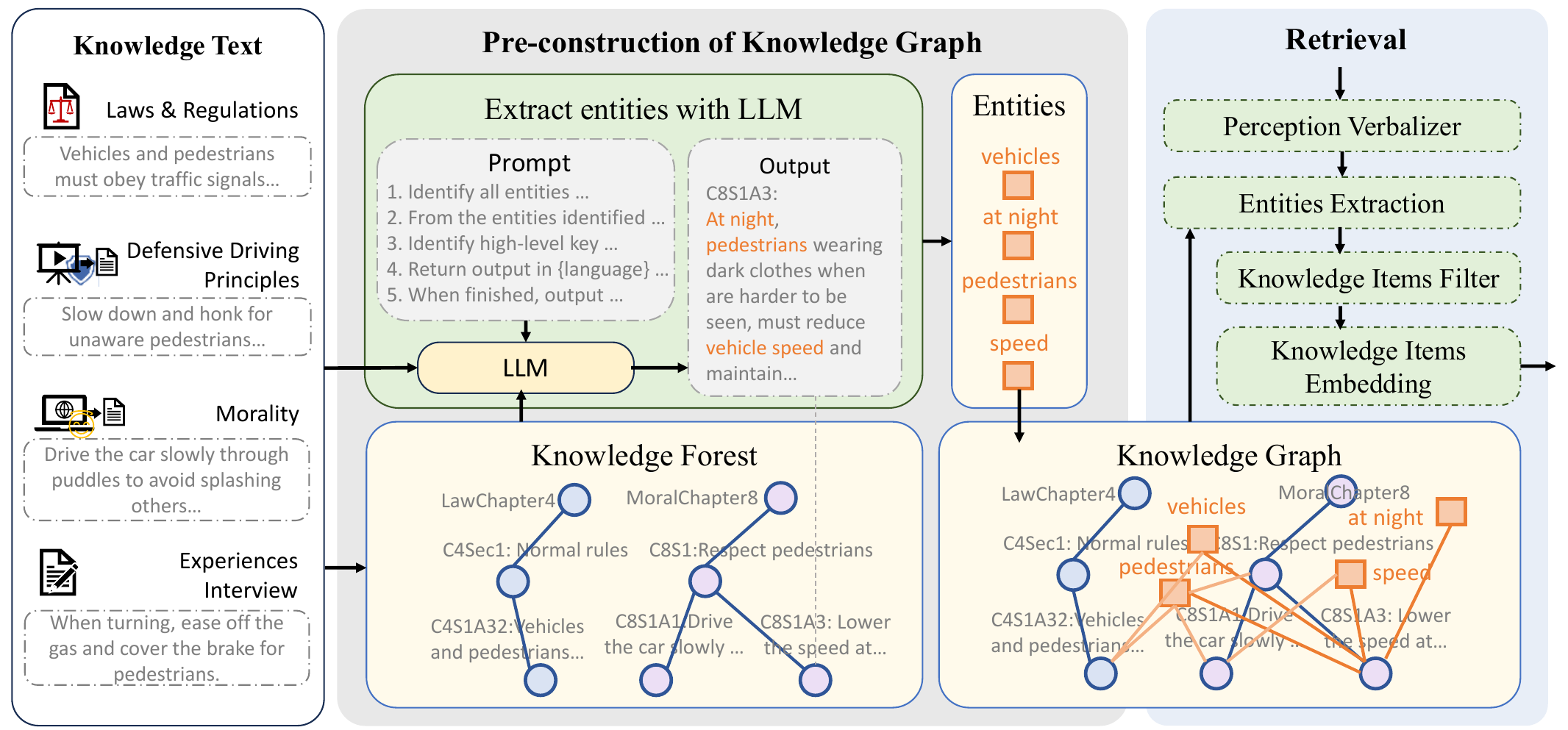}
    \vspace{-5pt}
\caption{\textbf{Pre-construction of the knowledge graph and retrieval process in \textsc{KnowVal}.}
We collect diverse driving-related resources—including laws, regulations, defensive driving principles, moral guidelines, and experiential knowledge—to construct an initial \textit{knowledge forest} based on textual structures. 
Large language models (LLMs) are then used to extract entities and define vertices and edges, forming a structured \textit{knowledge graph}. 
During inference, \textsc{KnowVal} generates queries enriched with 3D perception information to retrieve and rank relevant entries from the knowledge graph by descending relevance.
}
    \label{fig:graph}
    \vspace{-14pt}
\end{figure*}

\noindent \textbf{Planning with world prediction and value model.}
Current planning modules predominantly adopt Transformer or RNN architectures, which inherently possess varying degrees of capability for future-state prediction.
Building upon this foundation, we enhance both the planning and world models to generate diverse candidate trajectories and forecast corresponding future states.
We then introduce a Value Model that evaluates the desirability of these candidate trajectories and their predicted world states based on retrieved knowledge entries, as detailed in Section~\ref{sss:wm}.
To train this model, we construct a preference dataset comprising extensive trajectory–outcome samples, further described in Section~\ref{sss:data}.

\begin{figure*}[ht]
    \centering
    \includegraphics[width=0.88\textwidth]{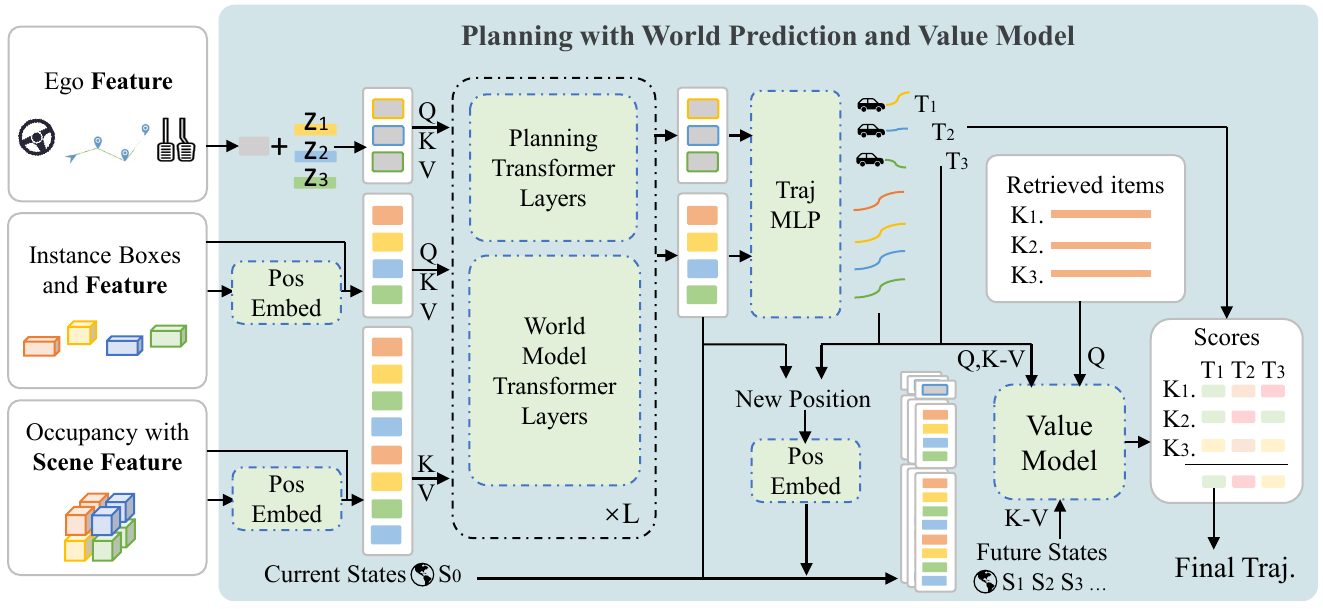}
    \vspace{-10pt}
\caption{\textbf{Planning process of \textsc{KnowVal}.}
We design a trajectory generation method compatible with existing Transformer- and RNN-based planners. 
By combining future world state prediction with value model assessment, \textsc{KnowVal} produces diverse candidate trajectories and makes interpretable final decisions.
}
    \label{fig:plan}
    \vspace{-14pt}
\end{figure*}

\subsection{Perception-guided Knowledge Retrieval.}


\subsubsection{Pre-Construction of Knowledge Graphs}
\label{sss:kg}
%
We design a knowledge graph construction method that preserves the complete fidelity of original knowledge clauses while simultaneously revealing latent semantic relationships between these clauses and query entities.

\noindent \textbf{Step 1: Knowledge collection and structured forest construction.}
We gather information from a variety of sources, including traffic laws and regulations, principles extracted from educational videos on defensive driving, transcripts from interviews with various drivers, and compiled ethical guidelines.
For educational videos on defensive driving, we employ Gemini 2.5-Pro with carefully designed prompts to extract key driving principles. The extracted texts are then categorized and organized into chapters according to the driving scenarios and strategies they describe.
For driver interview transcripts, we classify the content by driving behaviors and contexts (e.g., left turns, intersections) to produce a structured textual representation.
In the case of laws and regulations, their inherent hierarchical organization already exhibits a forest-like structure, which we directly adopt as the foundation for our knowledge forest.
This design allows us to directly utilize their intrinsic hierarchical organization to form the knowledge forest. 
Each knowledge clause node stores the raw, unaltered text, while non-leaf nodes contain the corresponding titles and descriptive information.
All nodes in the forest are labeled as \textit{native}.
This tree-based organization preserves the original hierarchical structure of the knowledge corpus and serves as the initial form of the knowledge graph.

\noindent \textbf{Step 2: Entity Linking via LLM to Transform the Forest into a Graph.}
To break down the “knowledge silos” among independent trees, we leverage LLMs to automatically discover and link semantic relationships across the disparate branches of the knowledge forest.
We iterate through each knowledge clause leaf node and employ a powerful LLM (e.g., GPT-4 or Qwen) to perform a controlled entity extraction task.
Our prompt instructs the model to identify key entities within the clause text corresponding to predefined conceptual categories, including:
\textit{Traffic-Sign-Device} (e.g., \textit{traffic signal}, \textit{stop sign}),
\textit{Road-User} (e.g., \textit{pedestrian}, \textit{bicycle}),
\textit{Driving-Maneuver} (e.g., \textit{turning}, \textit{merging}), and
\textit{Road-Condition} (e.g., \textit{wet pavement}, \textit{construction zone}).
After entity extraction, we create a new entity node in the knowledge graph for each unique entity and connect it to all corresponding \textit{native} nodes containing that entity.
The LLM then assigns semantic weights and relationship descriptions to the edges.
Inspired by~\cite{guo2024lightrag}, we further employ an LLM-empowered profiling function to generate text keys for both nodes and edges. Each key is a concise phrase facilitating efficient retrieval, and multiple keys can be generated to encode broader semantic themes across connected entities.
This process transforms the initially isolated knowledge forest into a highly interconnected, semantically rich knowledge graph, forming the structural backbone for the online retrieval mechanism described in Section~\ref{sss:retrieval}.

\subsubsection{Retrieval}
\label{sss:retrieval}
%
We design a retrieval process that ranks results by importance and relevance.
This process guarantees retrieval results faithful to the original text, thereby safeguarding against model hallucinations.

\noindent \textbf{Step 1: Perception verbalizer.} 
First, a templated verbalizer converts perception outputs of varying structures into a unified structured natural language.
For the 3D instances, the verbalizer directly converts the bounding box and semantic information into textual descriptions.
For the semantic map or occupancy results, the verbalizer first uses the Breadth-First Search algorithm to identify semantically connected blocks, handling discontinuous boundaries via a dynamic connectivity threshold. 
Upon obtaining connected blocks, it also converts their positional and semantic information into textual descriptions, which are then combined with abstract conceptual perception results, navigational data, and user instructions to form the retrieval query.
%

\noindent \textbf{Step 2: Entities Extraction.} 
%
We leverage a lightweight LLM to perform a two-layer keyword extraction from the retrieval query. 
The first layer extracts macro-level context keywords (e.g., \textit{Driving Security}, \textit{Scene Analysis}), while the second layer extracts specific entity and event keywords (e.g., \textit{Pedestrians}, \textit{Rainy Weather}). 
Subsequently, these keywords are used to query initial relevant entity nodes and edges within the knowledge graph by the index keys.
To capture richer contextual information, we perform a Top-K nearest-neighbor expansion on the graph, incorporating nodes closely connected to the initial set. 

\noindent \textbf{Step 3: Knowledge Items Filter and Embedding.} 
From all extracted nodes, we filter for only those explicitly tagged as \textit{native}. These nodes link directly to the original, verbatim knowledge clauses that have not been processed or summarized by any LLM. 
This filtering step is the fundamental differentiator of our system from conventional RAG, as it completely eliminates the risk of information distortion introduced by LLM processing.
%
The top-$N_K$ relevant filtered items are converted into vector representations in this process, making them suitable for subsequent use in the Value Model.
Meanwhile, the retrieval LLM outputs a list of items for supplementary perception, which are then fed back into the subsequent perception stage.


\begin{table*}[htbp]
    \centering
\caption{\textbf{Comparison of planning results on \textit{nuScenes}.}
L2 denotes the L2 error between the predicted and human driving trajectories.
}
    \vspace{-8pt}
    \label{tab:comparison_metrics_combined}
    \small 
    \resizebox{0.95\textwidth}{!}{
    \begin{tabular}{lc|cccc|cccl|cccc|cccl}
        \toprule
        \multirow{3}{*}{Method} & \multirow{3}{*}{Input} & \multicolumn{8}{c|}{UniAD Metrics} & \multicolumn{8}{c}{VAD/STP3 Metrics} \\
        \cmidrule(lr){3-10} \cmidrule(lr){11-18}
        & & \multicolumn{4}{c|}{L2 (m) $\downarrow$} & \multicolumn{4}{c|}{Collision Rate (\%) $\downarrow$} & \multicolumn{4}{c|}{L2 (m) $\downarrow$} & \multicolumn{4}{c}{Collision Rate (\%) $\downarrow$} \\
        \cmidrule(lr){3-6} \cmidrule(lr){7-10} \cmidrule(lr){11-14} \cmidrule(lr){15-18}
        & & 1s & 2s & 3s & Avg. & 1s & 2s & 3s & Avg. & 1s & 2s & 3s & Avg. & 1s & 2s & 3s & Avg. \\
        
        \midrule
        ST-P3~\cite{hu2022st}           & C & 1.72 & 3.26 & 4.86 & 3.28 & 0.44 & 1.08 & 3.01 & 1.51 & 1.33 & 2.11 & 2.90 & 2.11 & 0.23 & 0.62 & 1.27 & 0.71 \\
        \gr OccNet~\cite{tong2023scene}       & C & 1.29 & 2.13 & 2.99 & 2.14 & 0.21 & 0.59 & 1.37 & 0.72 & -&-&-&-&-&-&-&- \\
        UniAD~\cite{UniAD}                & C & 0.48 & 0.96 & 1.65 & 1.03 & 0.05 & 0.17 & 0.71 & 0.31 & 0.45 & 0.70 & 1.04 & 0.73 & 0.62 & 0.58 & 0.63 & 0.61 \\
        \gr VAD-Base~\cite{jiang2023vad}        & C & 0.54 & 1.15 & 1.98 & 1.22 & 0.04 & 0.39 & 1.17 & 0.53 & 0.41 & 0.70 & 1.05 & 0.72 & 0.07 & 0.17 & 0.41 & 0.22 \\
        PARA-Drive~\cite{weng2024drive} & C & -&-&-&-&-&-&-&- & 0.25 & 0.46 & 0.74 & 0.48 & 0.14 & 0.23 & 0.39 & 0.25 \\
        \gr SparseDrive~\cite{sun2025sparsedrive}   & C & 0.44 & 0.92 & 1.69 & 1.01 & 0.07 & 0.19 & 0.71 & 0.32 & 0.29 & 0.58 & 0.96 & 0.61 & 0.01 & 0.05 & 0.18 & 0.08 \\
        MomAD~\cite{song2025don}            & C & 0.43 & 0.88 & 1.62 & 0.98 & 0.06 & 0.16 & 0.68 & 0.30 & 0.31 & 0.57 & 0.91 & 0.60 & 0.01 & 0.05 & 0.22 & 0.09 \\
        \gr SSR~\cite{li2024navigation}                 & C & 0.24 & 0.65 & 1.36 & 0.75 & 0.00 & 0.10 & 0.36 & 0.15 & 0.18 & 0.36 & 0.63 & 0.39 & 0.01 & 0.04 & 0.12 & 0.06 \\
        BridgeAD-B~\cite{zhang2025bridging} & C & -&-&-&-&-&-&-&- & 0.28 & 0.55 & 0.92 & 0.58 & 0.00 & 0.04 & 0.20 & 0.08 \\
        \gr DiffusionDrive~\cite{liao2025diffusiondrive} & C & -&-&-&-&-&-&-&- & 0.27 & 0.54 & 0.90 & 0.57 & 0.03 & 0.05 & 0.16 & 0.08 \\
                
        \midrule
        GenAD~\cite{zheng2024genad}         & C & 0.36 & 0.83 & 1.56 & 0.91 & 0.06 & 0.23 & 1.00 & 0.43 & 0.28 & 0.49 & 0.78 & 0.52 & 0.08 & 0.14 & 0.34 & 0.19 \\
        \good KnowVal-GenAD & C & 0.46 & 1.35 & 2.72 & 1.51 & 0.00 & 0.20 & 0.80 & 0.33 \add{-0.10} & 0.28 & 0.60 & 1.07 & 0.65 & 0.01 & 0.08 & 0.25 & 0.11 \add{-0.08} \\

        \midrule
        HENet++~\cite{henetpp}      & RC & 0.39 & 1.11 & 2.36 & 1.29 & 0.00 & 0.06 & 0.33 & 0.13 & 0.24 & 0.50 & 0.91 & 0.55 & 0.01 & 0.03 & 0.10 & 0.05 \\
        \good KnowVal-HENet++ & RC & 0.42 & 1.19 & 2.35 & 1.32 & 0.00 & 0.10 & 0.25 & \textbf{0.12} \add{-0.01} & 0.26 & 0.54 & 0.94 & 0.58 & 0.00 & 0.01 & 0.08 & \textbf{0.03} \add{-0.02} \\
        \bottomrule
    \end{tabular}}
    \vspace{-14pt}
\end{table*}


\subsection{Planning with World Prediction and Value Model}
\label{ss:plan}

\subsubsection{Planning with World View and Values}
\label{sss:wm}

As illustrated in~\figref{fig:plan}, we first extend the planner to function as both a planning and world model, generating diverse trajectories and predicting future states.
Taking HENet++~\cite{henetpp} as an example, we inject $N_T$ distinct Gaussian noise vectors into the ego-vehicle features, thereby generating $N_T$ ego-specific queries that decode into $N_T$ distinct trajectories.
The baseline end-to-end model is fine-tuned with these modifications and an additional diversity constraint: when the L2 distance $e$ between two planned trajectories falls below a predefined threshold $\tau$, the term $(\tau - e)$ is added to the loss function as a penalty.
This encourages the planner to explore a wider range of feasible trajectories while maintaining consistency and realism.

Next, for each candidate trajectory $\mathcal{T}i$, \textsc{KnowVal} performs item-wise value assessment based on the retrieved knowledge entries ${K_j}$.
The Value Model follows a Transformer Encoder–MLP Decoder architecture.
For the $i$-th trajectory and the $j$-th knowledge entry feature $f_{K_j}\in \mathbb{R}^{N_\text{tokens}\times C}$, we take $max(f_{K_j})\in \mathbb{R}^C$ and concatenate it with the positional encoding of $\mathcal{T}i$ and its ego-vehicle features to form the query $Q_{i,j}$.
Through this process, $N_K$ query tokens are generated for each candidate trajectory.
These tokens are projected linearly and fed into a Transformer layer, while the future-state tokens $S_i$—which incorporate ego-vehicle, instance, and occupancy-grid features, each with a positional embedding corresponding to $\mathcal{T}i$—serve as key–value pairs.
After $L$ Transformer iterations, the updated $N_K$ features are decoded by the MLP to produce $N_K$ scalar values $s_{i,j} \in [-1, 1]$.
A value near $-1$ indicates behavior inconsistent with knowledge-based reasoning, $1$ represents positive compliance, and $0$ denotes irrelevance to the rule.

We then apply a weighted-decay scoring strategy to aggregate the per-rule evaluations into a single score for each trajectory.
The total value is defined as a normalized weighted average of scores across all retrieved rules, sorted by their relevance:
\begin{equation}
\text{Score}(\mathcal{T}i) = \frac{1}{Z} \sum_{j=1}^{N_K} \gamma^{j-1} \cdot s_{i,j}
\end{equation}
where $Z = \sum_{j=1}^{N_K} \gamma^{j-1}$ is a normalization constant.
This weighting ensures that the most relevant knowledge clauses contribute most significantly to decision-making, enabling interpretable, knowledge-driven trajectory evaluation.
Finally, the trajectory with the highest overall score is selected as the final planning output.

\subsubsection{Constructing Preference Dataset for Value Model}
\label{sss:data}
To train and evaluate the Value Model, we construct the Preference Dataset for Values, which provides quantitative supervision for the abstract notion of whether a driving behavior complies with a specific knowledge clause.
The dataset consists of 160K trajectory–knowledge pairs, each accompanied by corresponding scene states (feature vectors and BEV renderings) and ground-truth value annotations.
We use 80\% of the data for training and reserve the remaining 20\% for validation and testing.

\noindent \textbf{Step 1: Future States and Retrieved Basis.}
Our data construction pipeline begins by employing the retrieval engine described in Section~\ref{sss:retrieval} to obtain a set of relevant knowledge clauses $\text{Basis} = {K_1, K_2, \dots}$ for each driving scenario in nuScenes~\cite{nuScenes}.
We then collect the corresponding future-state vectors, BEV renderings of perception and prediction results, and other contextual information for each scenario.

\noindent \textbf{Step 2: Annotation Scores for Supervision.}
We design precise prompts and employ Qwen-VL-Max to automatically generate preliminary compliance scores for each data sample.
These scores serve as supervision signals, indicating how well each driving behavior aligns with the corresponding knowledge clauses.
To ensure annotation reliability, we conduct a manual review to identify and correct any apparent errors or outliers.
During training, the loss function is defined as the mean squared error (MSE) between the predicted value and the annotated score in the constructed dataset.

\section{Experiments}
\label{sec:exp}

\subsection{Datasets and Metrics}
We evaluate our approach on three widely used autonomous driving benchmarks: the real-world benchmark nuScenes~\cite{nuScenes} and NAVSIM~\cite{navsim}, and the simulation benchmark Bench2Drive~\cite{jia2024bench2drive}.
nuScenes~\cite{nuScenes} includes sensor data from six surround-view cameras, radar, and LiDAR across 1k driving scenes, about 35k samples.
Its open-loop planning metrics consist of the L2 error between predicted and human trajectories and the collision rate.
Bench2Drive~\cite{jia2024bench2drive} provides 10k sampled driving segments for training and offers closed-loop evaluation environments with detailed simulator configurations.
Bench2Drive employs two key metrics: Driving Score (combining Route Completion and Infraction Penalty) and Success Rate.
%
NAVSIM~\cite{navsim} introduces closed-loop metrics for open-loop evaluation, including the PDM score (PDMS) combining sub-metrics (NC, DAC, EP, TTC, C) over a 4-second non-reactive simulation. It includes 103k training and 12k testing samples.




\subsection{Main Results on Driving Benchmarks}
\label{sec:main_results}

%
%

\begin{table}[t]
\centering
\caption{\textbf{Closed-loop results on Bench2Drive.} 
DS denotes Driving Score, and SR denotes Success Rate (\%).}
\label{tab:my_updated_table}
\vspace{-8pt}
\resizebox{\columnwidth}{!}{
\begin{tabular}{lll|ll}
\toprule
Method & Expert & Paradigm & DS$\uparrow$ & SR$\uparrow$ \\
\midrule

AD-MLP~\cite{zhai2023rethinking} & Think2Drive & E2E & 18.05 & 00.00 \\ 
\gr TCP~\cite{wu2022trajectory} & Think2Drive & E2E & 40.70 & 15.00 \\ 
TCP-traj~\cite{wu2022trajectory} & Think2Drive   & E2E  & 59.90 & 30.00 \\
\gr UniAD-Base~\cite{UniAD} & Think2Drive   & E2E   & 45.81 & 16.36 \\
VAD~\cite{jiang2023vad}   & Think2Drive   & E2E   & 42.35 & 15.00 \\
\gr ThinkTwice~\cite{jia2023think} & Think2Drive & E2E & 62.44 & 31.23 \\ 
DriveAdapter~\cite{jia2023driveadapter} & Think2Drive & E2E & 64.22 & 33.08 \\
\gr DriveTrans.~\cite{jia2025drivetransformer}  & Think2Drive   & E2E   & 63.46 & 35.01 \\
ORION~\cite{fu2025orion}   & Think2Drive   & VLA  & 77.74 & 54.62 \\
\gr DiffusionDrive~\cite{liao2025diffusiondrive} & PDM-Lite & E2E  & 77.68 & 52.72 \\

\midrule
SimLingo~\cite{renz2025simlingo}  & PDM-Lite & VLA & 85.07 & 67.27 \\
\good KnowVal-SimLingo & PDM-Lite & KnowVal & \textbf{88.42} \add{+3.35} & \textbf{69.03} \add{+1.76} \\

\bottomrule
\end{tabular}
}
\vspace{-5pt}
\end{table}

\begin{table}[t]
\centering
\caption{\textbf{Open-loop results with closed-loop metrics on NVISIM.} }
\label{tab:navsim}
\vspace{-8pt}
\setlength{\tabcolsep}{2.5pt}
\resizebox{\columnwidth}{!}{
\begin{tabular}{l|c|ccccc|l}
\toprule

Method & Mod. & NC$\uparrow$ & DAC$\uparrow$ & TTC$\uparrow$ & Comf.$\uparrow$ & EP$\uparrow$ & \textbf{PDMS}$\uparrow$ \\
\midrule
VADv2~\cite{chen2024vadv2} & LC &  97.2 & 89.1 & 91.6 & 100 & 76.0 & 80.9 \\
\gr Hydra-MDP~\cite{li2024hydra} & LC & 98.3 & 96.0 & 94.6 & 100 & 78.7 & 86.5 \\
DiffusionDrive~\cite{liao2025diffusiondrive} & LC & 98.2 & 96.2 & 94.7 & 100 & 82.2 & 88.1 \\
\good KnowVal-DiffusionDrive & LC & 99.0 & 98.3 & 96.3 & 100 & 84.3 & 90.9 \add{+2.8} \\ 
\midrule
UniAD~\cite{UniAD} & C &  97.8 & 91.9 & 92.9 & 100 & 78.8 & 83.4 \\
\gr PARA-Drive~\cite{weng2024drive} & C & 97.9 & 92.4 & 93.0 & 99.8 & 79.3 & 84.0 \\
iPad~\cite{guo2025ipad} & C & 98.6 & 98.3 & 94.9 & 100 & 88.0 & 91.7 \\
\good KnowVal-iPad & C & 99.2 & 98.7 & 95.4 & 100 & 88.1 & 92.9 \add{+1.2} \\ 
\bottomrule
\end{tabular}
}
\vspace{-10pt}
\end{table}

\noindent \textbf{Open-loop Evaluation on nuScenes.}
As reported in~\tabref{tab:comparison_metrics_combined}, \ours~achieves the lowest Collision Rate among all existing methods on nuScenes.
Although the L2 distance is slightly higher, this primarily indicates deviations from human driving trajectories rather than degraded planning quality.
In practice, \ours~tends to discover safer or more efficient driving strategies that may not be reflected in human demonstrations.

\noindent \textbf{Closed-loop Evaluation on Bench2Drive.}  
As shown in~\tabref{tab:my_updated_table}, \ours~achieves a further \textbf{3.35} improvement in Driving Score and a \textbf{1.76\%} gain in Success Rate over the VLA-based method SimLingo, establishing new state-of-the-art performance on Bench2Drive.  

\noindent \textbf{Open-loop results with closed-loop metrics on NVISIM.}  
As shown in~\tabref{tab:navsim}, to further demonstrate the compatibility of \ours~with existing methods, we selected the advanced DiffusionDrive~\cite{liao2025diffusiondrive} and iPad~\cite{guo2025ipad} methods for experiments on NAVSIM. Both methods inherently support diverse candidate trajectory generation, thus eliminating the need for the modifications described in Section ~\ref{sss:wm}, and open-world perception, retrieval, and the value model can be directly integrated on them. 

For more implementation details, please refer to Appendix A.

Experiment results above demonstrate the effectiveness of our knowledge-grounded planning framework, and show that \ours~can be seamlessly integrated into existing end-to-end architectures, enhancing both safety and generalization.

\begin{figure*}[ht]
    \centering
    \includegraphics[width=\textwidth]{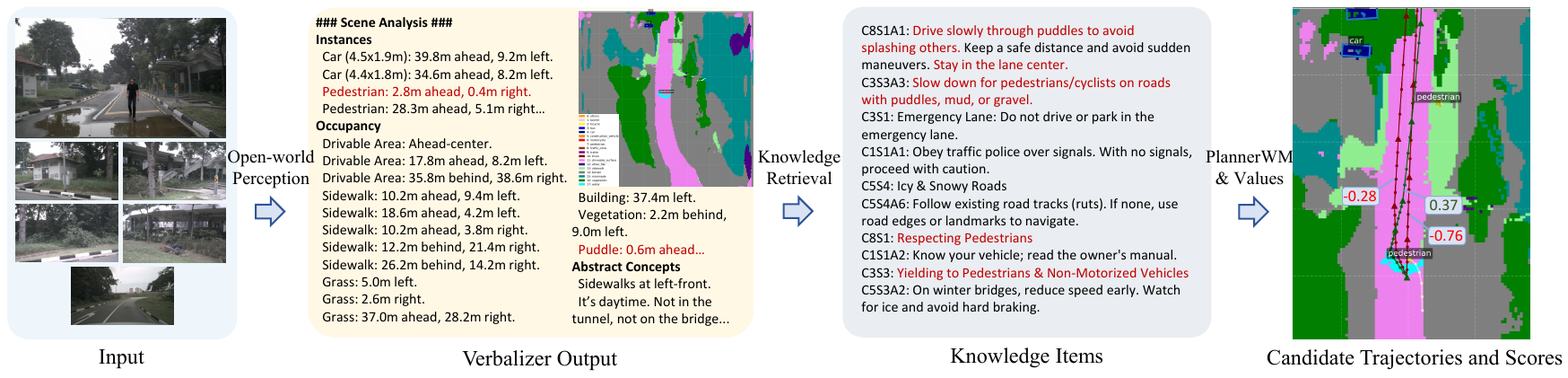}
    \vspace{-10pt}
    \caption{\textbf{Illustration of the inference process of \textsc{KnowVal}.} 
    \textsc{KnowVal} perceives long-tail scenarios, retrieves ethical knowledge related to pedestrians and puddles, and employs the Value Model to select appropriate actions.}
    \label{fig:sample}
    \vspace{-5pt}
\end{figure*}

\begin{figure}[ht]
    \centering
    \includegraphics[width=\columnwidth]{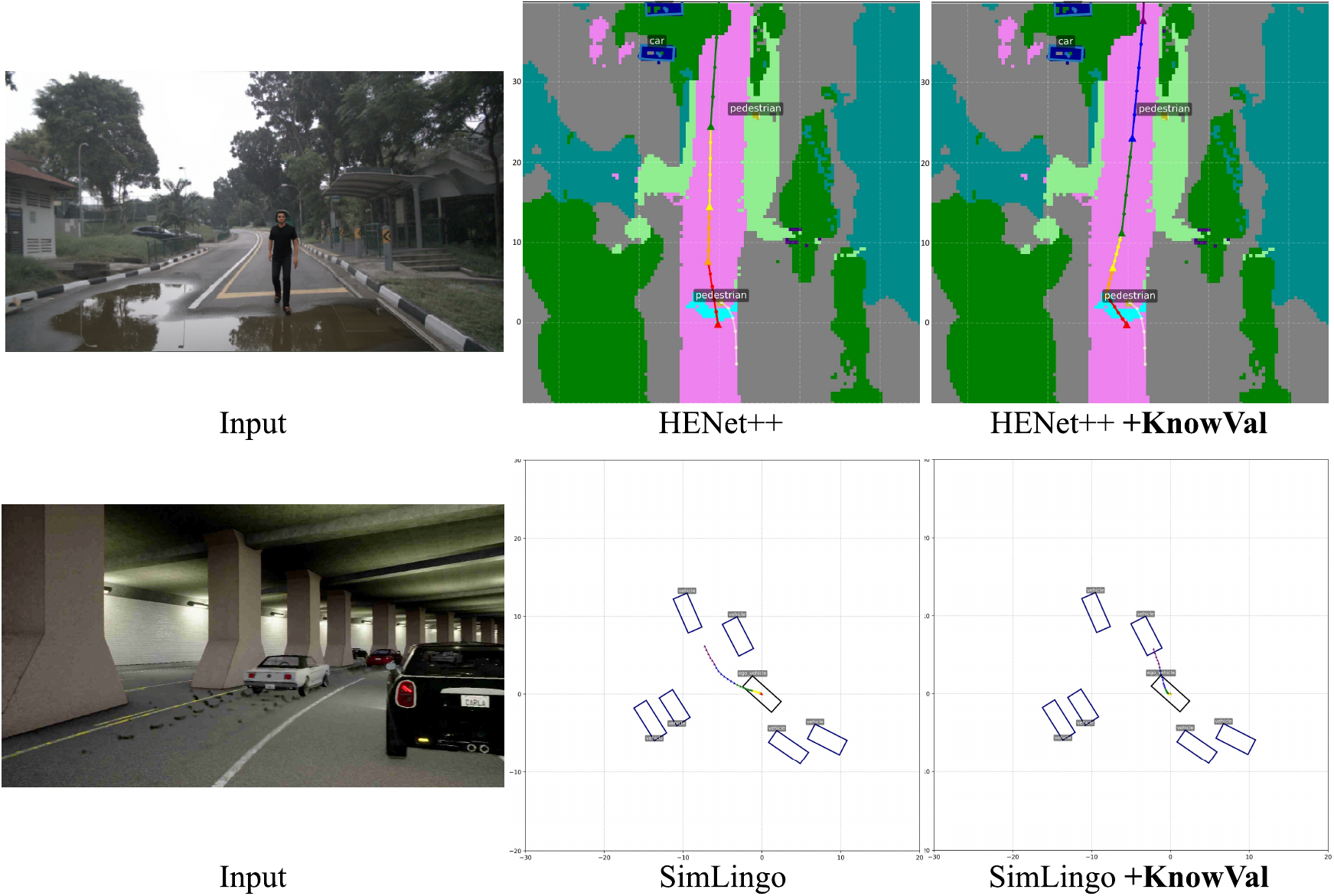}
    \vspace{-10pt}
    \caption{\textbf{Qualitative comparison between \textsc{KnowVal} and baseline methods.} 
    The upper example, edited following~\cite{song2025insert} on a nuScenes scenario, illustrates the \textbf{ethical commonsense}: “When passing through accumulated water, slow down to avoid splashing pedestrians.” 
    The lower example, sampled from Bench2Drive, demonstrates the \textbf{legal regulation}: “Inside tunnels, vehicles must not overtake across solid lines.” 
    Unlike baseline methods, \textsc{KnowVal} correctly handles both scenarios.}
    \label{fig:vis}
    \vspace{-10pt}
\end{figure}

\subsection{Qualitative Analysis}
\label{sec:qualitative}

\figref{fig:sample} illustrates a representative example of the inference process in our proposed \ours, highlighting its reasoning effectiveness.
Comparative results with baseline methods are presented in~\figref{fig:vis}.
Compared to existing approaches, our paradigm exhibits a superior ability to interpret and act upon ethical and legal considerations during decision-making.
It is important to note that current benchmarks lack both sufficient evaluation data and appropriate metrics to assess many forms of improper driving behavior.
In the examples shown in~\figref{fig:vis}, actions such as failing to yield to pedestrians or overtaking within tunnels do not affect standard metrics like collision rate or driving score, underscoring the limitations of existing evaluation protocols.

\subsection{Ablation}

As shown in~\tabref{tab:ablnusc} and~\tabref{tab:ablb2d}, integrating the knowledge retrieval and value model components, as well as incorporating open-world perception and retrieval-guided supplementary perception, consistently reduces the collision rate.
Although the deviation from human driving trajectories increases slightly, this does not imply a decline in planning quality.
We also analyze the effects of varying the numbers of knowledge entries and candidate trajectories in Appendix C. And we evaluate the robustness to noisy knowledge items in Appendix C.

\begin{table}[t]
\centering
\caption{
\textbf{Ablation studies on nuScenes.} 
The baseline model is GenAD. 
K\&V: using the proposed knowledge retrieval and value model. 
OW: incorporating open-ended perception and understanding of abstract concepts. 
RgP: using retrieval results to guide supplementary perception. 
}
\label{tab:ablnusc}
\vspace{-5pt}
\resizebox{0.65\columnwidth}{!}{
\begin{tabular}{ccc|cc}
\toprule
K\&V & OW & RgP & L2(m)$\downarrow$ & Col$\downarrow$ \\
\midrule

&  & &  0.91 & 0.43 \\
\gr \checkmark & &  & 0.89 & 0.38 \\
\checkmark & \checkmark &  & 1.47 & 0.34 \\ 
\good \checkmark & \checkmark & \checkmark  & 1.51 & 0.33 \\

\bottomrule
\end{tabular}
}
\vspace{-5pt}
\end{table}

\begin{table}[t]
\centering
\caption{\textbf{Ablation studies on Bench2Drive.}
The baseline model is SimLingo. 
OW: incorporating open-ended perception and understanding of abstract concepts. 
RgP: using retrieval results to guide supplementary perception.
}
\label{tab:ablb2d}
\vspace{-5pt}
\resizebox{\columnwidth}{!}{
\begin{tabular}{c|cccc|cc}
\toprule

ID & Knowledge & Values & OW & RgP   & DS$\uparrow$ & SR$\uparrow$ \\
\midrule
\gr A & &  &            &                  & 85.07 & 67.27 \\
B & \checkmark &  &            &                  & 84.91 & 67.03 \\
\gr C & & \checkmark &            &                  & 85.92 & 67.69 \\
D & & Heuristic Rules~\cite{sun2025sparsedrive} &            &                  & 85.94 & 67.82 \\
\gr E & \checkmark & \checkmark &            &                  & 87.61 & 68.40 \\
F & \checkmark & \checkmark & \checkmark &                  & 88.16 & 68.85 \\ 
\good G & \checkmark & \checkmark & \checkmark & \checkmark & 88.42 & 69.03 \\

\bottomrule
\end{tabular}
}
\vspace{-5pt}
\end{table}

\section{Conclusion}
\label{sec:conclusion}

This paper presents \ours, a new autonomous driving system that achieves visual–language reasoning through the synergistic interaction between open-world perception and knowledge retrieval.
Specifically, we construct a comprehensive knowledge graph for autonomous driving that integrates traffic laws, defensive driving principles, and ethical considerations, along with an efficient LLM-based retrieval mechanism tailored to driving contexts.
To enable value-aligned decision-making, we curate a human-preference dataset and train a Value Model that provides an interpretable foundation for trajectory evaluation.
Extensive experiments demonstrate that \ours~substantially enhances planning performance while maintaining full compatibility with existing methods, achieving the lowest collision rate on nuScenes and establishing state-of-the-art performance on Bench2Drive.
{
    \small
    \bibliographystyle{ieeenat_fullname}
    \bibliography{main}
}

\clearpage
\setcounter{page}{1}
\maketitlesupplementary

\renewcommand\thesection{\Alph{section}} 
\setcounter{section}{0} 

\lstset{
    basicstyle=\ttfamily\small, 
    breaklines=true,            
    frame=single,               
    backgroundcolor=\color{gray!10}, 
    columns=fullflexible,       
    keepspaces=true,            
    captionpos=b,               
    showstringspaces=false      
}

\section{Implementation Details}
\label{supp:detail}




\subsection{Training and Architecture Details}
\label{sec:implementation_details}

We employ Qwen2.5-3B for retrieval and knowledge embedding, with the number of retrieved knowledge entries set to $N_K=16$.
For the planner, the number of candidate trajectories is $N_T=20$, and the Value Model performs $L=3$ iterative reasoning steps.
The decay factor used in the total value assessment is $\gamma=0.7$.
To increase trajectory diversity, baseline models on nuScenes are fine-tuned for three epochs using eight A100 GPUs, while models on Bench2Drive are fine-tuned for one epoch under the same configuration.
The Value Model is trained separately on eight V100 GPUs for 50 epochs using the AdamW optimizer with a cosine annealing learning rate schedule.

\subsection{Prompt used for scoring trajectories with MLLM}

\begin{lstlisting}[caption={The prompt used for constructing GT of the Value Model Dataset}, label={lst:prompt1}]
# ROLE
You are an expert autonomous driving auditor. Your analysis is meticulous, objective, and follows a strict logical process. Your primary goal is to identify and quantify risk.

# TASK
For each of the prioritized driving rules provided below, you MUST follow a mandatory **4-Step Critical Evaluation Process**. Your final score for each rule is a direct consequence of this process. Do not skip any steps.

# PRIORITIZED RULES TO EVALUATE
{formatted_rules}

# MANDATORY 4-STEP CRITICAL EVALUATION PROCESS
For each rule, you must internally answer these questions in order:

**Step 1: Applicability Check**
*   **Question:** Is this rule relevant to the current scene? Can it possibly be applied or violated based on the visual context?
*   **Outcome:**
    *   If **NO** (e.g., a traffic light rule when no traffic lights are visible), the process stops here. Your conclusion is "Not Applicable". **The score MUST be 1.0.**
    *   If **YES**, proceed to Step 2.

**Step 2: Evidence Check**
*   **Question:** Does the image provide any visual information (positive or negative) to judge adherence to this applicable rule?
*   **Outcome:**
    *   If **NO** (e.g., a rule about checking blind spots, but the driver's actions are not visible), the process stops here. Your conclusion is "Applicable, but no evidence". **The score MUST be 0.0.**
    *   If **YES**, proceed to Step 3.

**Step 3: Adherence Check**
*   **Question:** Based on the available evidence, is the ego vehicle fully and correctly adhering to the rule?
*   **Outcome:**
    *   If **YES**, the process stops here. Your conclusion is "Perfect Adherence". **The score MUST be 1.0.**
    *   If **NO**, there is a violation. Proceed to Step 4 to determine its severity.

**Step 4: Risk Assessment (FOR VIOLATIONS ONLY)**
*   **Question:** How severe is the immediate risk created by this violation?
*   **Outcome:** Assign a risk level, which then determines the score range.
    *   **Negligible Risk:** A technical error with no immediate safety impact (e.g., stopping 10cm past the stop line with no cross-traffic). **Score: -0.1 to -0.2**.
    *   **Low Risk:** A poor driving habit or minor violation that increases potential risk but doesn't create immediate danger (e.g., late signaling in light traffic). **Score: -0.3 to -0.4**.
    *   **Moderate Risk:** A clear unsafe action that significantly increases risk (e.g., tailgating, changing lanes without signal near other cars). **Score: -0.5 to -0.7**.
    *   **High Risk:** A critical error that creates a probable chance of collision or puts others in immediate danger (e.g., failing to yield to a pedestrian entering a crosswalk, running a stale yellow light). **Score: -0.8 to -1.0**.

# CRITICAL INSTRUCTIONS
- **AVOID LAZY SCORING:** Do not default to 1.0 or 0.0 unless the 4-step process explicitly leads you there. The goal is to differentiate between different levels of non-compliance.
- **JUSTIFY YOUR SCORE:** The `conclusion` and the `risk_level` must logically and transparently lead to the final `score`.
- **BE SPECIFIC:** In your evidence description, refer to specific objects and actions in the scene.

# OUTPUT FORMAT
You MUST respond with a single JSON object. The object must have a single key "evaluations", which is a list. Each item in the list must adhere to the following structure:
1.  A "rule" key with the original rule text.
2.  A "reasoning" object containing four keys:
    - "positive_evidence": Visual cues that the rule is being followed. State "None" if none.
    - "negative_evidence": Visual cues that the rule is being violated. State "None" if none.
    - "risk_level": A string describing the risk for violations ("Negligible", "Low", "Moderate", "High"). For non-violations, state "N/A".
    - "conclusion": A summary of your 4-step analysis, explicitly stating the final judgment (e.g., "Not Applicable", "Perfect Adherence", "Moderate Risk Violation").
3.  A "score" key, a float number that is a direct consequence of your `conclusion` and `risk_level`.

Example of your required output format:
```json
{{
  "scene_description": "A vehicle is approaching an intersection with a pedestrian waiting to cross at a marked crosswalk. The traffic light is green.",
  "evaluations": [
    {{
      "rule": "Yield to pedestrians at the crosswalk.",
      "reasoning": {{
        "positive_evidence": "The vehicle is slowing down as it approaches the crosswalk.",
        "negative_evidence": "The vehicle does not come to a complete stop and proceeds through the crosswalk, forcing the pedestrian to wait.",
        "risk_level": "High",
        "conclusion": "High Risk Violation. The rule is applicable and evidence is clear. The vehicle failed to yield to a pedestrian ready to cross, creating immediate danger."
      }},
      "score": -0.8
    }},
    {{
      "rule": "Obey traffic signals.",
      "reasoning": {{
        "positive_evidence": "The traffic light for the vehicle's direction of travel is green, and the vehicle is proceeding through the intersection.",
        "negative_evidence": "None",
        "risk_level": "N/A",
        "conclusion": "Perfect Adherence. The rule is applicable, and the vehicle is correctly following the green light signal."
      }},
      "score": 1.0
    }},
    {{
      "rule": "No Honking.",
      "reasoning": {{
        "positive_evidence": "None",
        "negative_evidence": "None",
        "risk_level": "N/A",
        "conclusion": "Applicable, but no evidence. The rule is relevant in a city scene, but there is no visual or contextual information to determine if the horn was used."
      }},
      "score": 0.0
    }},
    {{
      "rule": "Stop at Railroad Crossing.",
      "reasoning": {{
        "positive_evidence": "None",
        "negative_evidence": "None",
        "risk_level": "N/A",
        "conclusion": "Not Applicable. There are no railway tracks or crossings visible in the scene."
      }},
      "score": 1.0
    }}
  ]
}}"""
\end{lstlisting}


\subsection{Prompt for keyword extraction in the retrieval stage}

\begin{lstlisting}[caption={The prompt used for retrieval}, label={lst:prompt2}]
---Goal---
Given a text document that is potentially relevant to this activity and a list of entity types, identify all entities of those types from the text and all relationships among the identified entities.
Use {language} as output language.

---Steps---
1. Identify all entities. For each identified entity, extract the following information:
- entity_name: Name of the entity, use same language as input text. If English, capitalized the name.
- entity_type: One of the following types: [{entity_types}]
- entity_description: Comprehensive description of the entity's attributes and activities
Format each entity as ("entity"{tuple_delimiter}<entity_name>{tuple_delimiter}<entity_type>{tuple_delimiter}<entity_description>)

2. From the entities identified in step 1, identify all pairs of (source_entity, target_entity) that are *clearly related* to each other.
For each pair of related entities, extract the following information:
- source_entity: name of the source entity, as identified in step 1
- target_entity: name of the target entity, as identified in step 1
- relationship_description: explanation as to why you think the source entity and the target entity are related to each other
- relationship_strength: a numeric score indicating strength of the relationship between the source entity and target entity
- relationship_keywords: one or more high-level key words that summarize the overarching nature of the relationship, focusing on concepts or themes rather than specific details
Format each relationship as ("relationship"{tuple_delimiter}<source_entity>{tuple_delimiter}<target_entity>{tuple_delimiter}<relationship_description>{tuple_delimiter}<relationship_keywords>{tuple_delimiter}<relationship_strength>)

3. Identify high-level key words that summarize the main concepts, themes, or topics of the entire text. These should capture the overarching ideas present in the document.
Format the content-level key words as ("content_keywords"{tuple_delimiter}<high_level_keywords>)

4. Return output in {language} as a single list of all the entities and relationships identified in steps 1 and 2. Use **{record_delimiter}** as the list delimiter.

5. When finished, output {completion_delimiter}

######################
---Examples---
######################
{examples}

#############################
---Real Data---
######################
Entity_types: [{entity_types}]
Text:
{input_text}
######################
Output:
\end{lstlisting}






\section{About the Knowledge Graph}
\label{supp:kg}

\begin{figure*}[b]
    \centering
    \includegraphics[width=\linewidth]{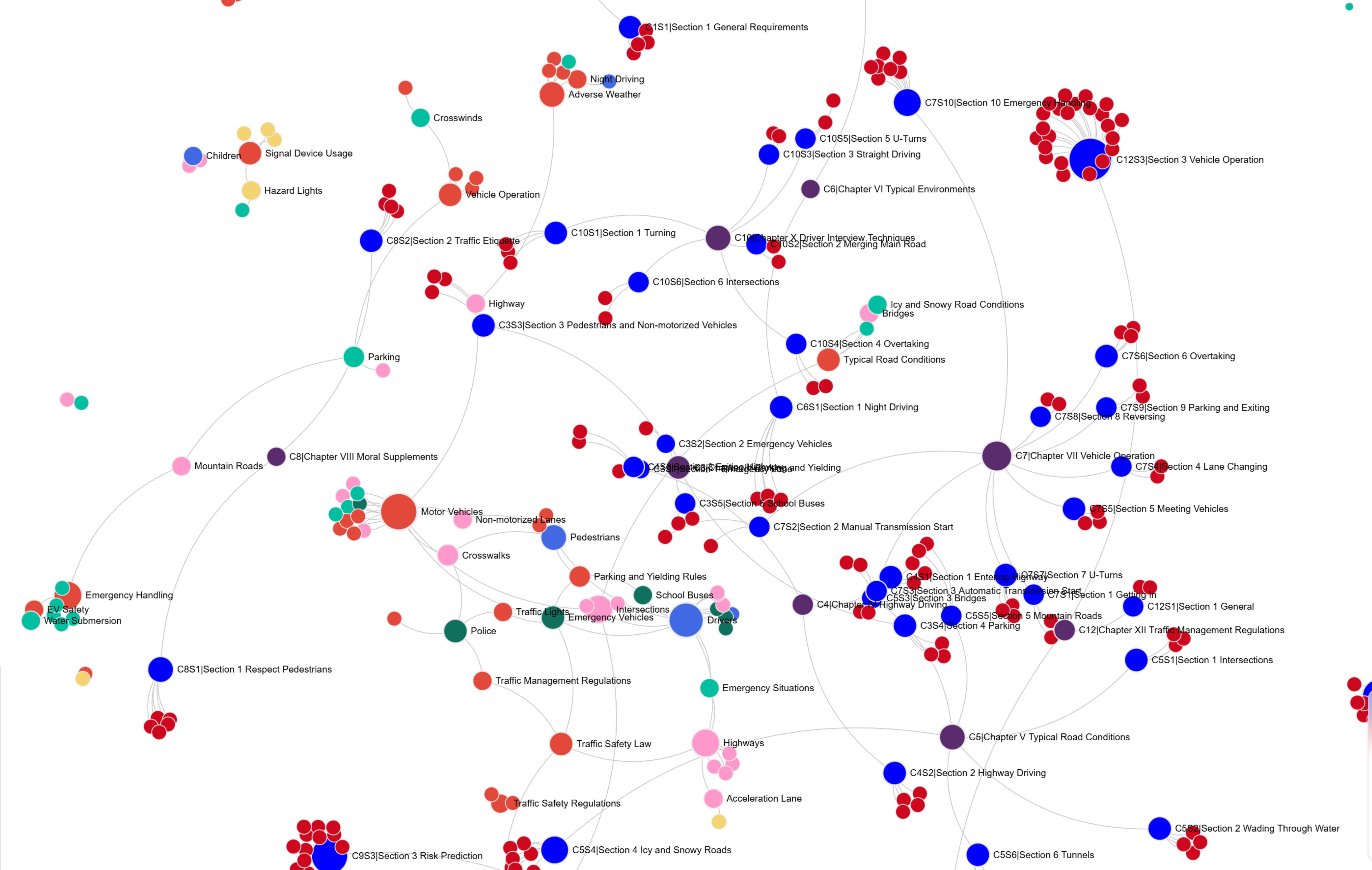}
    \caption{\textbf{Local view of the knowledge graph.}}
    \label{fig:kg}
\end{figure*}

A local view of the knowledge graph is shown in~\figref{fig:kg}.
Currently, the knowledge graph contains 1,324 nodes and 2,785 edges.

\section{Additional Ablation}
\label{supp:ablretri2}

\noindent \textbf{Ablation on nuScenes.}
As shown in~\tabref{tab:ablnusc2}, integrating the knowledge retrieval and value model components, as well as incorporating open-world perception and retrieval-guided supplementary perception, consistently reduces the collision rate.
Although the deviation from human driving trajectories increases slightly, this does not imply a decline in planning quality.
We further analyze the effects of varying the numbers of knowledge entries and candidate trajectories.
As shown in the middle and lower sections of the table, increasing these quantities beyond a certain threshold provides minimal additional benefit, indicating diminishing returns.

\noindent \textbf{Ablation on Bench2Drive.}
As shown in~\tabref{tab:ablb2d2}, The experiments show each component's effectiveness and robustness to noisy knowledge items.

\noindent \textbf{Ablation on the Value Model.}
As shown in~\tabref{tab:ablvalue}, we evaluate several architectures for the Value Model.
Using BEV images as input processed by CNNs offers the advantage of not requiring model-specific training for each baseline, but yields weaker predictive performance.
Introducing weighted averaging across disaggregated subscores effectively reduces prediction errors.
The entry labeled \textit{10K} represents our initial dataset version, which included only trajectories generated by HENet++.
In the final version, we augmented the dataset with numerous randomly generated low-quality trajectories as negative samples, significantly increasing both scale and diversity.

\begin{table}[t]
\centering
\caption{
\textbf{Ablation studies on nuScenes.} 
The baseline model is GenAD. 
K\&V: using the proposed knowledge retrieval and value model. 
OW: incorporating open-ended perception and understanding of abstract concepts. 
RgP: using retrieval results to guide supplementary perception. 
$N_K$: the number of retrieved knowledge entries. 
$W_K$: applying weighting to the total score.
$N_T$: the number of candidate trajectories.
}
\label{tab:ablnusc2}
\resizebox{\columnwidth}{!}{
\begin{tabular}{cccccc|cc}
\toprule
K\&V & OW & RgP & $N_K$ & $W_K$ & $N_T$ & L2(m)$\downarrow$ & Col$\downarrow$ \\
\midrule

&  & & 16 & \checkmark & 20 & 0.91 & 0.43 \\
\gr \checkmark & & & 16 & \checkmark & 20 & 0.89 & 0.38 \\
\checkmark & \checkmark & & 16 & \checkmark & 20 & 1.47 & 0.34 \\ 
\good \checkmark & \checkmark & \checkmark & 16 & \checkmark & 20 & 1.51 & 0.33 \\

\midrule
\checkmark & \checkmark & \checkmark & 16 & \checkmark & 10 & 1.56 & 0.37 \\
\good \checkmark & \checkmark & \checkmark & 16 & \checkmark & 20 & 1.51 & 0.33 \\
\checkmark & \checkmark & \checkmark & 16 & \checkmark & 30 & 1.52 & 0.33 \\

\midrule
\gr \checkmark & \checkmark & \checkmark & 8 & \checkmark & 20 & 1.29 & 0.36 \\ 
\checkmark & \checkmark & \checkmark & 16 & & 20 & 1.67 & 0.38 \\
\good \checkmark & \checkmark & \checkmark & 16 & \checkmark & 20 & 1.51 & 0.33 \\
\checkmark & \checkmark & \checkmark & 24 & \checkmark & 20 & 1.49 & 0.34 \\ 

\bottomrule
\end{tabular}
}
\end{table}

\begin{table}[h]
\centering
\caption{\textbf{Ablation studies on Bench2Drive.}
Exp.~A is the SimLingo baseline.
Exp.~B uses the retrieved language tokens as Key-Value pairs for the planning transformer.
Exp.~C trains the value model using only the future-state tokens $S_i$ as input—excluding knowledge entries—while keeping the training setting consistent.
Exp.~D uses heuristic rules (SparseDrive; Sun et al., 2025) rather than a learned value model.
Exp.~H randomly inserts $N_k$ random knowledge items into the retrieved entries, while Exp.~I uses $N_k$ randomly selected knowledge items without any retrieval.
The experiments show each component's effectiveness and robustness to noisy knowledge items.
}
\label{tab:ablb2d2}
\resizebox{\columnwidth}{!}{
\begin{tabular}{c|cccc|cc}
\toprule

ID & Knowledge & Values & OW & RgP   & DS & SR \\
\midrule
\gr A & &  &            &                  & 85.07 & 67.27 \\
B & \checkmark &  &            &                  & 84.91 & 67.03 \\
\gr C & & \checkmark &            &                  & 85.92 & 67.69 \\
D & & Heuristic Rules &            &                  & 85.94 & 67.82 \\
\gr E & \checkmark & \checkmark &            &                  & 87.61 & 68.40 \\
F & \checkmark & \checkmark & \checkmark &                  & 88.16 & 68.85 \\ 
\good G & \checkmark & \checkmark & \checkmark & \checkmark & 88.42 & 69.03 \\

\midrule
H & Noise      & \checkmark &            &                  & 87.07 & 67.99 \\
\gr I &  Random     & \checkmark &            &                  & 86.18 & 67.74 \\
J & \checkmark & \checkmark &            &                  & 87.61 & 68.40 \\

\bottomrule
\end{tabular}
}
\end{table}

\begin{table}[t]
\centering
\caption{\textbf{Comparison on different Value Model architectures.}}
\label{tab:ablvalue}
\resizebox{\columnwidth}{!}{
\begin{tabular}{llc|cc}
\toprule
Input $S_i$ & Output & Traning Scale & MSE$\downarrow$ &  MAE$\downarrow$ \\
\midrule
BEV Image & Total Score & 10K & 0.35 & 0.59 \\
\gr BEV Image & Total Score & 128K & 0.22 & 0.44 \\
Scene Feature & Total Score & 128K & 0.13 & 0.27 \\
\good Scene Feature & Subscores & 128K & 0.11 & 0.20 \\

\bottomrule
\end{tabular}
}
\end{table}

\section{Analysis of Retrieval Validity}
\label{supp:ablretri2}
We conducted statistics on nuScenes \textit{val}. \textbf{50.1\%} of the retrieved items are Key Items (which certainly fit in this scenario), \textbf{41.8\%} of them are General Items (fit in any scenario, e.g., speed limits), and \textbf{8.1\%} of them are Irrelevant Items (e.g., `give way to ambulance', but there is no ambulance). 
Crucially, the fraction of Irrelevant Items did not exceed 2/16 in any single trajectory.
\textbf{The rows H to J in~\tabref{tab:ablb2d2} also demonstrate robustness to noisy entries}, as we train the value model to score 0 for irrelevant entries.

\section{Qualitative Analysis of the Value Model}
\label{supp:vm}



To better illustrate the role of our proposed and trained Value Model,~\figref{fig:visvm} showcases several examples and visualizes ego vehicle trajectories across two scenarios, where trajectories 2 and 4 were manually constructed as \textit{negative samples} and the black trajectory in the center represents the ego vehicle.
In the first case (top), the first trajectory successfully avoids pedestrians and the leading vehicle, whereas the trajectory in the second image heads directly toward them. A comparison of the predicted scores for knowledge entries 1, 3, and 5 demonstrates that the Value Model has successfully learned the concepts of "maintaining a safe following distance" and "respecting and yielding to pedestrians." 

In the second scenario (bottom, images 3 and 4), a large truck is present ahead of the ego vehicle. According to defensive driving principles, one should maintain a distance from large, risky vehicles. The comparison of predicted scores for knowledge entries 1 and 4 indicates that the Value Model has effectively learned the concepts of "maintaining a safe following distance" and "avoiding large vehicles."
In both scenarios, the scores predicted by our Value Model closely align with the GT. 

The scores predicted by our value model in both scenarios align closely with the ground truth. This demonstrates the model's ability to effectively gauge knowledge compliance across various contexts and assign reasonable scores accordingly.

\begin{figure*}[th]
    \centering
    \includegraphics[width=\textwidth]{img/visvm.pdf}
    \caption{\textbf{Visualization of future state evaluation results using the proposed Value Model.} Displayed are BEV of the future state, retrieved knowledge entries, and Prediction vs. GT Scores. The comparison between compliant (Rows 1, 3) and risky (Rows 2, 4) trajectories demonstrates that our Value Model can evaluate future states against knowledge.}
    \label{fig:visvm}
\end{figure*}

\section{Sample of the Value Model Dataset}
\label{supp:data}

%
Some samples from the value model dataset are shown in~\figref{fig:visvm}.
%
We will make the entire dataset and the model publicly available in the future.

\end{document}